\documentclass[times, review, 10pt]{elsarticle}
\usepackage{graphicx} 
\usepackage{epstopdf}
\usepackage[export]{adjustbox} 
\usepackage{float}
\usepackage{amsmath}
\usepackage{amssymb}
\usepackage{tabularx}
\usepackage{booktabs}
\usepackage{array}
\usepackage{setspace}
\usepackage{tocloft}
\usepackage{color}
\usepackage{microtype}
\usepackage[a4paper, left=1.91cm, right=1.91cm, top=2.54cm, bottom=2.54cm]{geometry}
\usepackage[colorlinks,
            linkcolor=blue,
            anchorcolor=blue,
            citecolor=green, 
            urlcolor=blue]{hyperref}

\newcolumntype{C}[1]{>{\centering\arraybackslash}p{#1}}

\begin{document}

\begin{frontmatter}

\title{Efficient Neural Implicit Representation for 3D Human Reconstruction}

\author[1]{Zexu Huang}
\author[1]{Sarah Monazam Erfani}
\author[1]{Siying Lu}
\author[2]{Mingming Gong\corref{cor1}}

\address[1]{School of Computing and Information Systems, The University of Melbourne}
\address[2]{School of Mathematics and Statistics, The University of Melbourne}

\cortext[cor1]{Corresponding author}

\begin{abstract}
High-fidelity digital human representations are increasingly in demand in the digital world, particularly for interactive telepresence, AR/VR, 3D graphics, and the rapidly evolving metaverse. Even though they work well in small spaces, conventional methods for reconstructing 3D human motion frequently require the use of expensive hardware and have high processing costs. This study presents HumanAvatar, an innovative approach that efficiently reconstructs precise human avatars from monocular video sources. At the core of our methodology, we integrate the pre-trained HuMoR, a model celebrated for its proficiency in human motion estimation. This is adeptly fused with the cutting-edge neural radiance field technology, Instant-NGP, and the state-of-the-art articulated model, Fast-SNARF, to enhance the reconstruction fidelity and speed. By combining these two technologies, a system is created that can render quickly and effectively while also providing estimation of human pose parameters that are unmatched in accuracy. We have enhanced our system with an advanced posture-sensitive space reduction technique, which optimally balances rendering quality with computational efficiency. In our detailed experimental analysis using both artificial and real-world monocular videos, we establish the advanced performance of our approach. HumanAvatar consistently equals or surpasses contemporary leading-edge reconstruction techniques in quality. Furthermore, it achieves these complex reconstructions in minutes, a fraction of the time typically required by existing methods. Our models achieve a training speed that is 110X faster than that of State-of-The-Art (SoTA) NeRF-based models. Our technique performs noticeably better than SoTA dynamic human NeRF methods if given an identical runtime limit. HumanAvatar can provide effective visuals after only 30 seconds of training. Please visit \href{https://github.com/HZXu-526/Human-Avatar}{https://github.com/HZXu-526/Human-Avatar} for further demo results and code.
\end{abstract}


\end{frontmatter}
\begin{figure}
    \vspace{-0.5cm}
    \centering
    \includegraphics[width=1\textwidth,height=0.28\textwidth]{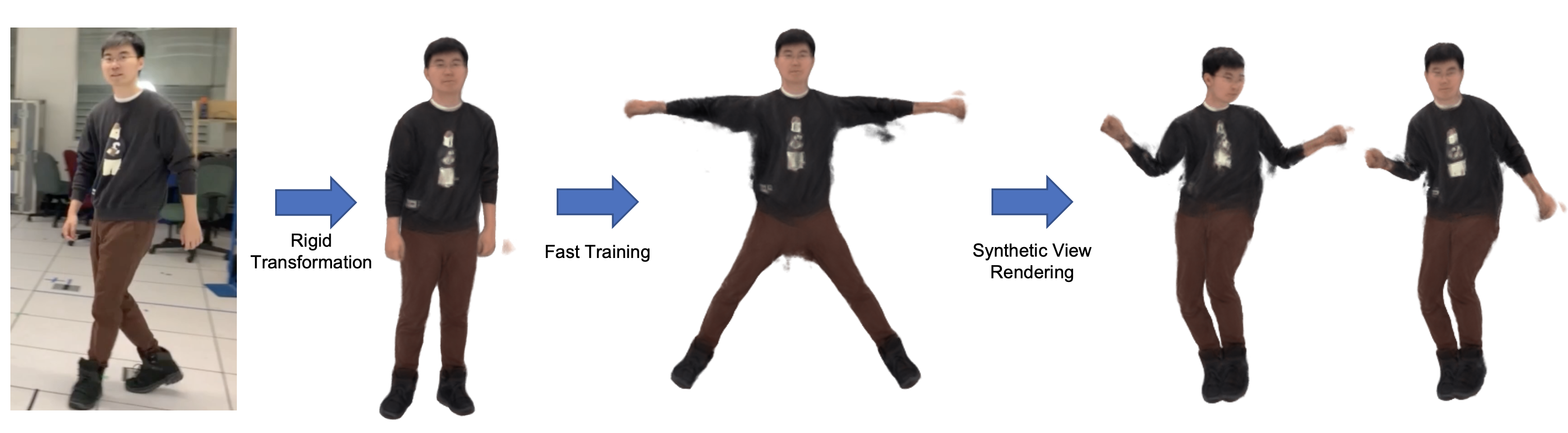}
    \caption{HumanAvatar: We present a framework that produces computationally realistic human avatar representations from single-camera video footage, incorporating both poses and facial features. After reconstruction, the avatar can be animated and displayed at a rate of 15 frames per second with a resolution of 540x540 pixels. To accomplish our goals, we combine a pre-trained model specialized in human motion with efficient neural radiance fields, which were initially created for static environments. We also incorporate a rapid articulation correspondence search mechanism. By leveraging an established technique for skipping empty spaces, we enhance both the training and inference speeds, making it possible to learn avatar dynamics within minutes.}
    \label{fig:intro}
\end{figure}

\section{Introduction}
Numerous applications, such as interactive telepresence, AR/VR, 3D graphics, and the developing metaverse depend on the ability to create high-fidelity digital humans. Existing techniques such as multiple synchronized cameras~\cite{shysheya2019textured}, RGB-D sensor~\cite{li2020robust}, and magnetic trackers~\cite{state1996superior} are the most effective for capturing 3D human motion within a restricted space. However, these existing methods require expensive instruments to capture motion and accumulate high processing costs, making them inaccessible to most researchers. Given a single RGB video as input, it is desirable to reconstruct realistic human motion and acquire a 3D human mesh. The scarcity and difficulty of obtaining 3D motion datasets underscore the need to expand these resources for training motion generation models, such as ``text to motion'' models~\cite{tevet2023human, shafir2023human}. However, previous 3D human reconstruction methods~\cite{saito2019pifu, alldieck2018detailed} still struggle to accurately capture details like hair and textile creases. In this paper, we set out to develop a lightweight, broadly deployable system for learning smoother 3D virtual people from monocular video independently which is sufficiently quick to support real-world situations. (see Fig.~\ref{fig:intro}). \par

In the field of computer vision, models that capture the essential geometry, texture, and appearance to accurately depict human movement are crucial. In recent literature, a multitude of approaches have emerged. The HuMoR model~\cite{rempe2021humor}, a notable auto-regressive human motion estimation method, excels in predicting plausible shapes and poses from unclear data sources. Owing to its training on a diverse motion capture dataset, HuMoR is adept at generalizing across a range of movements and body types, processing inputs from 3D keypoints, and RGB or RGB-D videos. However, its application is confined to pose estimation and it does not support the generation of novel movements beyond its trained repertoire. \par

Recent advances focus on motion scene reconstruction through neural rendering, which combines principles of physics with deep learning in computer graphics. Neural networks can now capture the intricacies of complex scenes, producing photorealistic outputs. A testament to this progress is the introduction of Neural Radiance Fields (NeRF)~\cite{mildenhall2021nerf}, which have demonstrated proficiency in understanding 3D scenes and objects. NeRF can synthesize realistic images from novel viewpoints by processing 5D coordinates, including spatial locations (x, y, z) and viewing directions ($\theta$, $\phi$), and translating them into volume density and colour. Over the past two years, various models have expanded upon NeRF to specifically render dynamic humans~\cite{jiang2022neuman, su2021nerf, peng2021animatable, chen2021animatable, peng2021neural, liu2021neural}, leading to the emergence of Dynamic Human NeRFs. These methods typically represent the human figure and appearance in a pose-invariant canonical space. They use animation and rendering techniques like skinning algorithms to morph and display the model in posed space, facilitating reconstruction from images of humans in various postures. This interplay between posed and canonical spaces allows for network optimization by minimizing the discrepancy between reconstructed pixel values and actual images. Dynamic Human NeRFs also employ human parametric models like SMPL~\cite{loper2015smpl} for frame-by-frame animation, using NeRF's rendering capabilities to recreate human movements. A significant challenge with these methods is the lack of consistency in the 3D human models across frames, which can lead to inaccuracies. Moreover, the extensive training and inference times required by these models present practical deployment challenges.\par

In this paper, we intend to develop a novel Dynamic Human NeRF model that can reconstruct smoother human motion efficiently. In order to achieve this, we propose HumanAvatar, an approach that estimates a smooth human pose parameter and reconstructs high-fidelity avatars from a monocular video within a few minutes. The avatar can be animated and generated at dynamic speeds after it has been trained. Accomplishing such a speedup and smooth estimation is a difficult endeavour that needs meticulous method design, speedy differentiable rendering and effective implementation. The utility of the Dynamic Human NeRF model extends significantly into real-world applications, offering a more convenient and expedited means of integrating high-fidelity human avatars into AR and VR environments. This capability not only enhances user engagement by providing immersive and interactive experiences but also facilitates a variety of practical applications ranging from virtual meetings and remote learning to interactive gaming and virtual social interactions. \par

Our straightforward effective pipeline includes different essential elements. First, we leverage a newly proposed pre-trained human motion model HuMoR~\cite{rempe2021humor} to estimate more accurate human pose parameters. HuMoR~\cite{rempe2021humor} estimates future human posture parameters based on past and present human posture parameters. Since it incorporates correlations between video frames, the estimation is smoother than when correlations between frames are not included. Second, we use a newly invented variation of the neural radiance field~\cite{muller2022instant} for learning the canonical pose and shape. Through the use of a more effective hash table in place of multi-layer perceptrons (MLPs), Instant-NGP~\cite{muller2022instant} speeds up the process of volume rendering. However, Instant-NGP is only capable of handling rigid objects since the spatial properties are explicitly specified. Third, we integrate the conventional NeRF with a powerful articulation model Fast-SNARF~\cite{chen2023fast} that allows learning from posed inputs and the capacity to move the avatar. Fast-SNARF~\cite{chen2023fast} is hundreds of times quicker than its slower version, SNARF~\cite{chen2021snarf}, and effectively generates a constant deformation field to bend the canonical space into the pose field. \par

However, merely merging current acceleration methodologies will not produce the needed efficiency. As the quick articulation section and acceleration mechanisms for the canonical field are in place, rendering the real volume turns into the processing bottleneck. Normal volume rendering requires querying and accumulating the density of hundreds of locations along the light ray in order to determine the colour of a pixel. The practice of keeping the occupancy grid to skip pixels in the empty area is a typical way to speed up this process. Nevertheless, such a method presupposes static situations and cannot be used for dynamic applications especially moving humans. \par

To address this challenge, we apply a posture-sensitive space reduction to form articulated structures in dynamic settings. Our method gathers pixel points from a standard grid within the pose field for each input pose during inference, repositioning these samples to the canonical pose for density calculations. We then use these densities to craft a spatial occupancy grid that bypasses empty spaces in volumetric rendering. For training, we maintain a persistent occupancy grid over all frames, tracking occupied regions across time. The parameters of this grid are periodically updated with densities from randomly sampled points across various training frames, balancing computational efficiency with rendering quality. \par

To assess the efficacy of our approach, we conduct evaluations using both computer-generated and real-world one-camera videos featuring humans in motion. We make comparisons with the latest innovative techniques developed specifically for the reconstruction of monocular avatars. Remarkably, our methodology not only matches the quality of reconstructions achieved by these state-of-the-art methods but also outperforms them in terms of animation fidelity. The integration of HuMoR, Instant-NGP, and Fast-SNARF represents a significant innovation in our work. It offers a synergistic effect that elevates the capabilities of our Dynamic Human NeRF model beyond the sum of its parts. Additionally, we offer a thorough ablation analysis to better understand how various elements within our system contribute to its speed and precision. \par

\subsection*{Contributions}
In summary, the contributions of this work are as follows:

\begin{enumerate}
    \item We integrate a pre-trained human motion model into a dynamic human NeRF in the preprocessing procedure, which estimates more accurate human parameters and reduces the jitter of the predicted human parameters to reconstruct higher-quality human avatars.

    \item We propose an accelerated framework that replaces the basic volume rendering method in the dynamic human NeRF with a more efficient rendering method and integrates an efficient articulation human model from SoTA for transforming the human body from canonical space to pose space.

    \item We integrate a posture-sensitive space reduction approach in the process of volumetric rendering in our framework. This strategy maintains an equilibrium between rendering quality and the effectiveness of computation. It can further speed up our inference and training speed.

    \item We conduct experiments using real and synthetic data to compare the accuracy of human reconstruction. In comparison to approaches that estimate human pose parameters in each frame independently, results show that the estimated human pose parameters are better matched to humans. Moreover, our reconstruction speed is much faster than methods using NeRF.
    
\end{enumerate}

\noindent 
The remainder of this paper is organized as follows: Section 2 details the related work, while Section 3 describes our methodology. Section 4 discusses the experimental setup, model configurations, performance comparisons of our approach against other state-of-the-art (SOTA) dynamic human NeRFs, ablation studies and limitations. Finally, Section 5 concludes with a summary of our contributions.
\section{Related Work}
\noindent \textbf{3D Reconstruction on Humans}\ \ \ \
The challenge of accurately rendering the three-dimensional form and appearance of humans has been an enduring area of research. Previously, exceptional levels of reconstruction have been accomplished as demonstrated in studies~\cite{collet2015high, dou2016fusion4d, guo2019relightables}, utilizing an extensive assembly of cameras or depth-capturing devices. However, the high costs associated with such equipment confine their use to professional environments. More recent studies~\cite{alldieck2018detailed, alldieck2018video, guo2021human, habermann2019livecap, xu2018monoperfcap} have shown the potential for reconstructing 3D humans from single-camera video footage by using either bespoke or standard template mesh models like SMPL~\cite{loper2015smpl}. These approaches achieve 3D form by adapting the template to align with two-dimensional joint positions and outlines. Nevertheless, tailored mesh templates may not always be obtainable, and standard mesh templates often fall short in replicating intricate details and varying styles of apparel.

Neural representations~\cite{mescheder2019occupancy, oechsle2019texture, mildenhall2021nerf} have recently been recognized for their robustness in modelling three-dimensional humans~\cite{bergman2022generative, chen2021animatable, chen2022gdna, chen2021snarf, chibane2020implicit, liu2021neural, peng2021animatable, peng2021neural, saito2019pifu, saito2020pifuhd}. These techniques have enabled various studies~\cite{chen2021animatable, guo2023vid2avatar, jiang2022neuman, liu2021neural, peng2021animatable, peng2021neural, jiang2023instantavatar} to accurately create detailed neural-based human avatars by using a limited number of images or even a single-camera video, eliminating the need for a pre-scanned individualized template. Furthermore, the recently proposed MonoHuman~\cite{yu2023monohuman} is to obtain a canonical NeRF (Neural Radiance Field) ~\cite{mildenhall2021nerf}representation, which is accomplished by warping camera rays from the observation space into the canonical space. This is done in order to obtain density and radiance values straight from the canonical NeRF. Although these methods produce high-quality results and can reconstruct avatars from monocular video data, their effectiveness is limited by slow training and processing times. This limitation stems from the inherent latency involved in canonical representation and deformation computations. Our proposed method tackles this bottleneck, making it feasible to learn avatars in a matter of minutes.

\vspace{1em}

\noindent\textbf{Rendering Radiance Field}\ \ \ \
Numerous strategies have been advanced to enhance the speed of training and inference for neural representations~\cite{chen2022tensorf, garbin2021fastnerf, liu2020neural, muller2022instant, fridovich2022plenoxels, yu2021plenoctrees}, focusing on substituting multilayer perceptrons (MLPs) with more expedient formats within these representations. Some studies~\cite{liu2020neural, fridovich2022plenoxels, yu2021plenoctrees} have suggested employing voxel grids for depicting neural fields, resulting in accelerated training and inference. Instant-NGP~\cite{muller2022instant} takes this a step further by utilizing a multi-resolution hash table in place of dense voxels, which is not only more space-saving but also capable of capturing finer details. Additionally, to enhance rendering efficiency, other studies~\cite{liu2020neural, muller2022instant} have focused on using an occupancy grid to bypass voids. These approaches significantly speed up both training and inference processes.

\vspace{1em}

\noindent
\textbf{3D Gaussians}\ \ \ \
3D Gaussian Splatting~\cite{kerbl20233d} is recently proposed and it is defined as a point-based technique for scene representation that enables high-quality real-time rendering. This method of scene rendering can be expressed using a collection of 3D Gaussians. The initial 3D Gaussians are mostly concerned with static images. As 3D Gaussian Splatting (3DGS)~\cite{kerbl20233d} is highly effective in terms of quality and speed, a wide range of research works have explored the use of the 3D Gaussian model for dynamic scene reconstruction. As a result, 3D Gaussian-based avatar reconstruction\cite{hu2024gauhuman, saito2024relightable, qian20243dgs, yuan2024gavatar, zielonka2023drivable, ye2023animatable} has rapidly evolved, quickly establishing itself as a dynamic research domain. These methods utilize the strength of 3D Gaussians for reconstructing avatars but often come with drawbacks such as irregular clustering and initial bias. Consequently, avatars developed through this method tend to exhibit prominent artifacts during novel pose animations. The D3GA~\cite{zielonka2023drivable} proposes to incorporate 3D Gaussians into tetrahedral frameworks and then use these framework modifications to animate avatars. Nevertheless, it deviates from our original goal by requiring an extra 3D scan in order to fabricate the base three-dimensional mesh structure and depending on inputs obtained from densely calibrated multi-perspective videos. Ye et al.~\cite{ye2023animatable} employ rigid body movements and pose-dependent deformations to adapt 3D Gaussians inside a canonical framework. However, this method requires two hours of training time and does not present results based on inputs from a single camera. Similarly, Li et al.~\cite{li2024animatablegaussians} applied 2D CNNs to improve radiance field renderings during the post-processing phase in order to create avatars with intricate visuals based on the analysis of multi-view video data. However, there are rendering speed limitations with this approach.

Despite the remarkable quality and efficiency these methods offer for rigid body objects, adapting them to non-rigid bodies presents challenges. Our approach balances quality and training speed and merges the capabilities of Instant-NGP with a novel articulation algorithm to facilitate animation and learning from posed data. Additionally, we introduce a novel concept for posture-sensitive space reduction, specifically designed for animated, articulated human avatars.
\section{Methodology}
\begin{figure}
    \vspace{-0.5cm}
    \centering
    \includegraphics[width=1\textwidth]{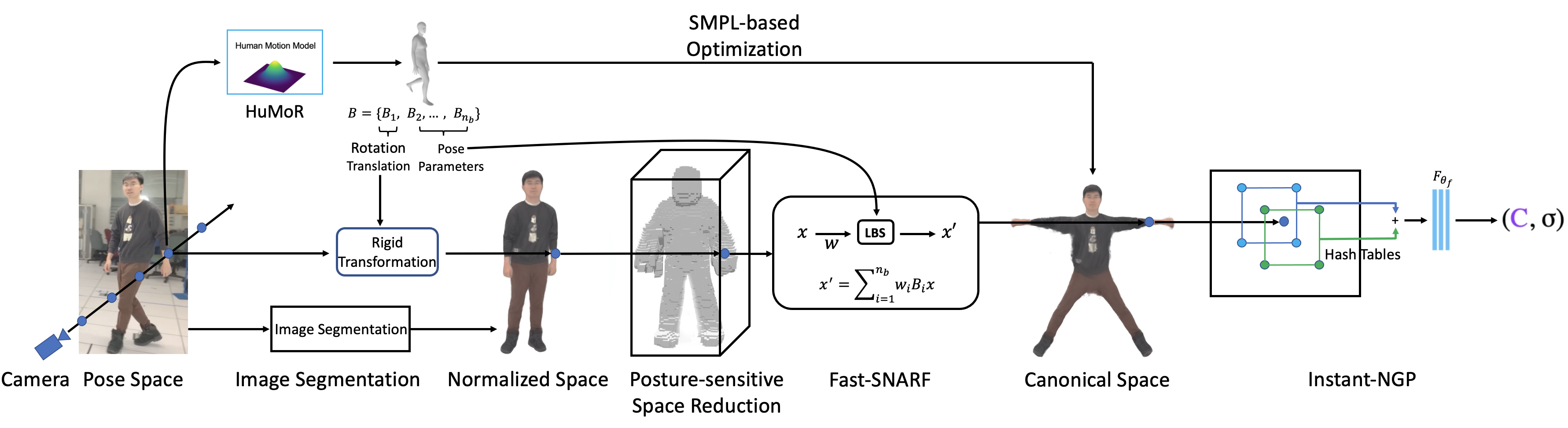}
    \caption{\textbf{Model Structure:} We estimated the SMPL parameters using HuMoR for each frame and sampled the rays' pose space positions. The global orientation and translation are then subtracted from these points' locations in a normalised space, after which our occupancy vector is used to filter out points in unoccupied space. We use the SMPL body model to optimize the human avatar in the canonical space. To assess the colour and density attributes, additional data points are incorporated into a canonical neural radiance field. This is achieved by utilizing an articulation module that warps these points into the canonical space.}
    \label{fig:model_structure}
\end{figure}
Our main goal is to properly reconstruct a 3D avatar of the moving human subject in a video. In this section, we discuss the deployment of a segmentation model to generate human masks. After that, we elaborate on our unique framework, which integrates the human motion model and the volumetric rendering method. This integration effectively transforms radiance fields into observable space. Additionally, our framework includes an accelerated neural radiance field, engineered to capture both the physical form and aesthetic attributes within a canonical space, along with a dedicated articulation module. To circumvent the unnecessary sampling of voids within the 3D bounding box encompassing the human body, which is predominantly composed of empty space, our approach introduces a specialized posture-sensitive space reduction strategy tailored for humans. Finally, we delve into the training objectives that guide our model's learning process and the regularization strategies.
\subsection{Image Segmentation}
Our preprocessing suite is encapsulated within the Segmentation-Anything Model (SAM)~\cite{kirillov2023segment} which sets a new standard in image segmentation. SAM's architecture melds a pre-trained Vision Transformer (ViT) with a versatile prompt encoder and an efficient mask decoder, establishing it as a robust solution for segmentation tasks. This targeted approach helps our model to train the human part. Furthermore, this segmentation precision plays a crucial role in the training phase, especially in optimizing the loss function, ensuring that our model concentrates on human figures, thereby sharpening the accuracy of our estimations. Within our architectural framework, the SAM is utilized to distinguish the human subject from the background, effectively extracting the figure while excluding non-relevant elements. Throughout the model's training iterations, these generated human masks are instrumental in refining our algorithm, being integral to the loss function computation. This is achieved by using the masks to focus the comparison between the actual human pixel values and those reconstructed by the model, thereby enhancing the precision of our human avatar reconstruction.
\subsection{Human Motion Model}
In our framework, we leverage a SoTA pre-trained human motion model, HuMoR (Human Motion Model for Robust Estimation)~\cite{rempe2021humor}, which is one of the time-series human motion models to estimate the SMPL body mesh~\cite{loper2015smpl} of the human in the monocular video. \par

\vspace{1em}

\noindent\textbf{State Representation}\ \ \ \ 
It captures the underlying dynamics of human motion through a state representation matrix $\textbf{X}$. This matrix comprises root translation $r$, root orientation $\Phi$, body pose joint angles $\Theta$, and joint positions $J$, mathematically expressed as 
\begin{equation}
    \textbf{X} = \begin{bmatrix}
    r & \dot{r} & \Phi & \dot{\Phi} & \Theta & J & \dot{J}
    \end{bmatrix},
\end{equation}
where $\dot{r}, \dot{\Phi},$ and $\dot{J}$ are the representation of the root and joint velocities. The model's latent variable dynamics focus on modelling the probability of a time sequence of states, which is crucial for capturing the nuances of human motion. The latent space can be interpreted as generalized forces, serving as inputs to a dynamics model with numerical integration. \par

\vspace{1em}

\noindent\textbf{Latent Variable Dynamics Model}\ \ \ \
Despite not being explicitly linked to physical laws, the latent space of this model can be thought of as generalised forces. These generalised forces are the inputs of a numerically integrated dynamics model. This can be expressed as:
\begin{equation}
    \textbf{X}_T = f(\textbf{X}_0, z_{1:T}),
\end{equation}
where $\textbf{X}_T$ is the state at time $T$, $\textbf{X}_0$ is the initial state, and $z_{1:T}$ is the sequence of latent transitions. HuMoR incorporates this latent dynamics model when predicting human poses, which gives the motion temporal information. This allows the video to predict situations where the human body is slightly occluded. \par

\vspace{1em}

\noindent\textbf{Test-time Motion Optimization}\ \ \ \
Test-time Motion Optimization (TestOpt) operates in synergy with the HuMoR model to facilitate the recovery of human posture and form from observational data that may be occluded or contain errors. Given a temporal sequence of observations $y_{0:T}$, TestOpt aims to deduce the underlying body shape $\beta$ and a series of SMPL pose parameters $(r_{0:T}, \Phi_{0:T}, \Theta_{0:T})$. This optimization process is particularly noteworthy for its use of the CVAE's initial state $x_0$ and a sequence of latent transitions $z_{1:T}$ to parameterize the motion. \par

In the preprocessing stage of our model, we employ the Test-time Motion Optimization procedure of the pre-trained HuMoR model to split our input video into different frames, and then estimate SMPL pose parameters for the person in this video. We use a 6D transformation matrix $B$ to represent HuMoR's output $\textbf{X}$ in the stage of rigid transformation and canonical space transformation, which can be expressed as:
\begin{equation}
    \textbf{X} = B = \{B_1, \ldots, B_{n_b}\}.
\end{equation}
We employed the Fast-SNARF~\cite{chen2023fast}, where we use the Linear Blend Skinning (LBS) algorithm to compute coordinates in the canonical space, shown in Eq. 7. This algorithm necessitates the use of the bone transformations $B = \{B_1, \ldots, B_{n_b}\}$ to calculate the transformed coordinate. After obtaining the coordinates, we apply the Instant-NGP~\cite{muller2022instant} to calculate colour and volume density for these coordinates within the canonical space. The sampling required for the radiance field, symbolized as \(F_{\theta_f}\), is also accomplished through the human body parameters, which can be shown as:
\begin{equation}
    F_{\theta_f}(x^*, T(B)) = C,
\end{equation}
where $C$ is the pixel colour in the canonical space, $x^*$ is the canonical coordinates of the human and $T$ represents the process of transforming the matrix $B$ into the input for the Rendering Radiance Field. 

In the rendering stage of the human body within our canonical space, we utilize the SMPL state representation matrix $\textbf{X}$ obtained through HuMoR as an occupancy grid to optimize the quality of the human avatar in the canonical space. This approach allows us to eliminate extraneous pixels rendered beyond a certain distance from the pose parameters in the state representation matrix $\textbf{X}$. Consequently, it facilitates the successful training of an avatar that represents a human figure.
\subsection{Rendering Radiance Field}
Our method employs the articulated radiance fields to synthesize new viewpoints. We trace a ray for each pixel from the camera origin \(\mathbf{o}\) through the pixel in direction \(\mathbf{d}\), expressed as \(\mathbf{r} = \mathbf{o} + t\mathbf{d}\). We sample \(N\) points along this ray within the scene's depth bounds, transforming these points back into their canonical positions before evaluating them with our foundational radiance field \(F_{\theta_f}\), depicted in Figure \ref{fig:model_structure}. This process retrieves the colour and density for each point on the ray in the articulated field \(F'_{\theta_f}\). We calculate the pixel color \(C\) by accumulating the products of the color \(c_i\) and the transmittance \(\alpha_i\) along the ray, using the equation:
\begin{equation}
    C = \sum_{i=1}^{N} \alpha_i \left( \prod_{j<i} (1 - \alpha_j) \right) c_i, \quad \text{where} \quad \alpha_i = 1 - \exp(-\sigma_i \delta_i),
\end{equation}
with \(\delta_i\) representing the distance between consecutive samples. Our research targets this by optimizing neural rendering for dynamic human avatars, prioritizing the efficient exclusion of empty space. This optimization balances performance with output fidelity, contributing to the advancement of neural rendering techniques.

\vspace{1em}

\noindent\textbf{Efficient Neural Radiance Field for Canonical Space}\ \ \ \
We utilize a radiance field, symbolized as $F_{\theta_f}$, to characterize both the shape and appearance of a human figure in a standard canonical space. This radiance field is designed to estimate the density $\sigma$ and color $c$ for every 3D coordinate $x$ within this canonical area, expressed as:
\begin{equation}
    F_{\theta_f} : \mathbb{R}^3 \to \mathbb{R}^3, \mathbb{R}^+
\end{equation}
\begin{equation}
    x \mapsto c, \sigma,
\end{equation}
where $\theta_f$ are the parameters that define the radiance field. To parameterize $F_{\theta_f}$, we employ the Instant-NGP~\cite{muller2022instant} technique. This method uses a hash table to hold feature grids at several granularity levels, enabling quick training and inference. Instant-NGP uses a tri-linear interpolation of the characteristics at neighbouring grid points while attempting to predict the textural and geometric qualities of a query spatial location. A shallow Multi-Layer Perceptron (MLP) is used to aggregate and finally decode these interpolated characteristics that were gathered from various levels.

\vspace{1em}

\noindent\textbf{Articulating Radiance Fields}\ \ \ \
In our model, generating animations and facilitating learning from images in specific poses requires the production of deformed radiance fields in the target poses, denoted as $F'_{\theta_f}$. This pose space is represented by a radiance field which can be described by the following function:
\begin{equation}
    F'_{\theta_f}: \mathbb{R}^3 \rightarrow \mathbb{R}^3, \mathbb{R}^+
\end{equation}
\begin{equation}
    x' \mapsto \sigma, c.
\end{equation}
Here, $x'$ maps to both the colour and density in the posed space. In order to simulate articulation, we employ a skinning weight space $s$ in canonical space, where $\sigma_s$ are its parameters:
\begin{equation}
    S_{\sigma_s}: \mathbb{R}^3 \rightarrow \mathbb{R}^{n_b},
\end{equation}
\begin{equation}
    x \mapsto s_1, \ldots, s_{n_b},
\end{equation}
that $n_b$ stands for the total number of bones in the skeletal structure. Fast-SNARF~\cite{chen2023fast} renders this skinning weight field as a low-resolution voxel grid to avoid the computing expense of SNARF~\cite{chen2021snarf}. The skinning parameters of each grid point's closest vertex on the SMPL body model~\cite{loper2015smpl} are used to calculate its value. The point $x$ in the canonical field is changed to deformed space $x'$ by linear blend skinning using the canonical skinning space and objective bone transformations $B = \{B_1, \ldots, B_{n_b}\}$ as shown below:
\begin{equation}
    x' = \sum_{i=1}^{n_b} S_i B_i x.
\end{equation}
With the inverse mapping of Equation 7, the canonical coordinates $x^*$ for a warped point $x'$ are identified. The most important step is to define the mapping between points in the pose space $x'$ and the corresponding points in the canonical space $x^*$. Fast-SNARF's root-finding method is an effective way to accomplish this. Consequently, the presented radiance field $F'_{\theta_f}$ may be written as:
\begin{equation}
    F'_{\theta_f} (x') = F_{\theta_f} (x^*),
\end{equation}
where $F_{\theta_f}$ is the radiance field that we use.
\subsection{Posture-sensitive Space Reduction}
We observe that the human anatomy, due to its articulated nature, is mostly comprised of empty spaces within the three-dimensional bounding volume that surrounds it. This leads to substantial empty spaces within the bounding volume. This characteristic significantly hinders the rendering process by necessitating numerous redundant sampling operations. While precomputed occupancy grids provide a partial solution for static models, they fall short when applied to dynamic humans, where the voids shift with the movements, making the rendering task more complex.

\begin{figure}
    \centering
    \includegraphics[width=1\textwidth,height=0.3\textwidth]{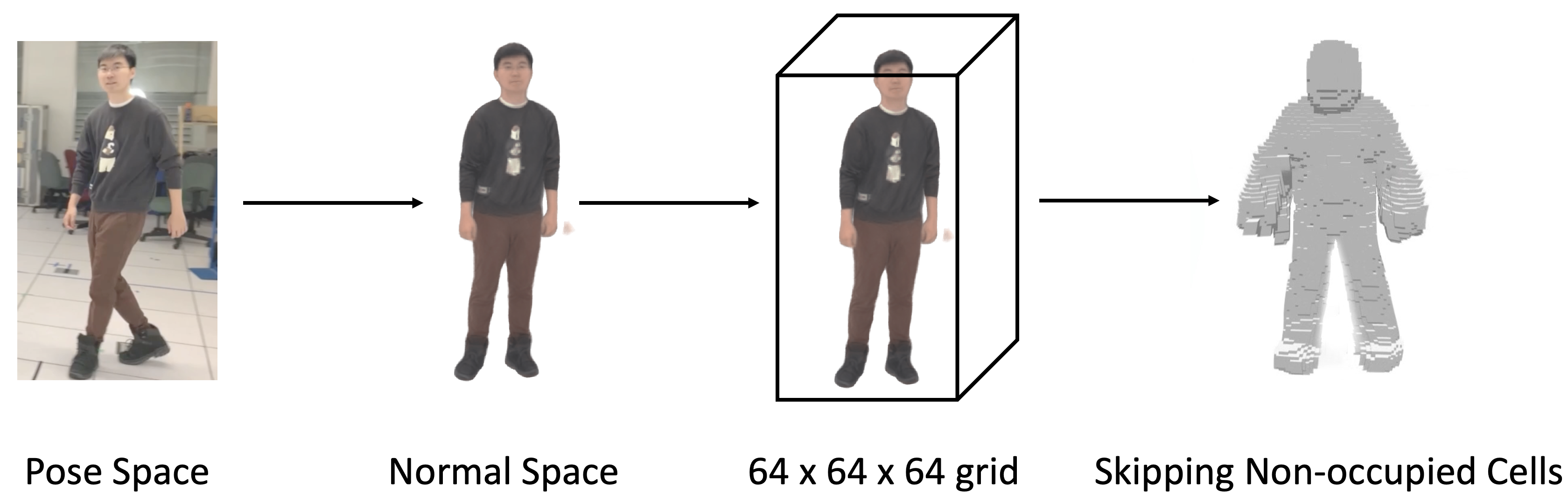}
    \caption{\textbf{Posture-sensitive Space Reduction Procedure.} In the inference stage, our method skips redundant sampling.}
    \label{fig:PSSR}
\end{figure}

\vspace{1em}

\noindent \textbf{Inference Phase:} During inference, our method entails generating a set of points on a \(64 \times 64 \times 64\) grid corresponding to the posed space for each given body posture as shown in the Fig. \ref{fig:PSSR}. We query these points against our posed radiance field \(F_{\theta_f}\) to obtain their densities, which we then convert to binary occupancy values. To fully encapsulate the subject and reduce errors from the grid's coarse resolution, we use a dilation technique to expand the occupancy regions, thus preventing potential false negatives. Given the grid's low granularity and the vast number of queries required for comprehensive rendering, the creation of this occupancy grid imposes a negligible computational burden. In the volumetric rendering process, we enhance efficiency by setting the density of unoccupied cells to zero. This eliminates the need for further queries to\(F_{\theta_f}\) and streamlines the rendering operation.

\vspace{1em}

\noindent \textbf{Training Phase:} During training, the computational burden accumulates with each iteration due to the creation of an occupancy grid. To address this, we adopt a more holistic strategy by initializing a unified occupancy grid at the onset of training, which aggregates the occupied regions from multiple frames. This grid is updated regularly every \(k\) iteration, leveraging the moving average of current occupancy against the densities queried from the posed radiance field \(F_{\theta_f}\). Notably, this grid operates within a normalized coordinate field, which strips away the human's translation and global orientation. This minimizes the volume of occupied space, thereby reducing redundant queries and enhancing computational efficiency.
\subsection{Training Losses}
We fine-tune our model by employing a robust variant of the Huber loss, the Huber loss is defined as:
\begin{equation}
    \text{Huber}(y, f(x)) = 
    \begin{cases} 
    \frac{1}{2}(y - f(x))^2 & \text{if } |y - f(x)| \leq \delta \\
    \delta |y - f(x)| - \frac{1}{2}\delta^2 & \text{otherwise},
    \end{cases}
\end{equation}
where $y$ is the true value, $f(x)$ is the predicted value, and $\delta$ is a user-defined threshold parameter. We symbolized the Huber loss as $\rho$, to quantify the divergence between the estimated pixel colour $C$ and its corresponding ground truth pixel colour $C_{\text{gt}}$:
\begin{equation}
    L_{\text{rgb}} = \rho(\| C - C_{\text{gt}} \|).
\end{equation}
Additionally, we introduced our approach of image segmentation in Section 3.1, which contains the segmentation of people and contexts with the Segment Anything Model (SAM)~\cite{kirillov2023segment}. Among them, the segmented people are retained by us as mask information. In the training phase, we will use this mask information to optimize our model. We proceed under the presumption that a rough estimation of the human mask is available. Then, in order to reduce the presence of spatial artefacts, we apply a specialised loss function on the values of the 2D alpha channel:
\begin{equation}
    L_{\alpha} = \| \alpha - \alpha_{\text{gt}} \|_1.
\end{equation}
For the purpose of Hard Surface Regularization, we follow the methodology~\cite{rebain2022lolnerf} to introduce supplementary regularization terms. These terms are designed to guide the NeRF model towards the prediction of more physically plausible surfaces:
\begin{equation}
    L_{\text{hard}} = -\log(\exp^{-|\alpha|} + \exp^{-|\alpha - 1|}) + \text{const.}
\end{equation}
where const. is a constant term that is used in this case to guarantee that the number of loss do not stray into the negative range. By enabling the early termination of rays whenever the cumulative opacity reaches a value of one, this type of regularisation helps to speed up the rendering process. \par

In the realm of Occupancy-based Regularization, the SMPL body model has been used as a regularizer in previous techniques~\cite{jiang2022neuman, chen2021animatable} for the learning of human avatars, which frequently encourage models to anticipate zero density for locations outside of the surface and solid density for sites inside the surface. By doing this, artefacts close to the body's surface are reduced. However, such regularization can be inadequate for loosely fitting clothing due to over-reliance on the body's shape assumptions and has been found lacking in removing artifacts close to the body. To address this, we forgo the use of the SMPL model for regularization and instead introduce an additional loss function, $L_{\text{reg}}$, which encourages points within the unoccupied cells of our occupancy grid to adopt zero-density. This approach allows for a more conservative estimation of both the subject's form and that of their clothing. The loss $L_{\text{reg}}$ is in the form of:
\begin{equation}
    L_{\text{reg}} = 
    \begin{cases} 
    |\sigma(x)|& \text{if \( x \) resides in empty space} \\
    0 & \text{otherwise}.
    \end{cases}
\end{equation}
\section{Experiments}
Our evaluation encompasses both accuracy and speed assessments of our approach using monocular video inputs. We benchmark our method against current state-of-the-art (SoTA) alternatives. Moreover, we conduct an ablation study to dissect the impact of each technical component within our methodology.
\subsection{Implementation Details}
Our approach was developed in Python, utilizing the PyTorch library, and involved the creation of custom CUDA kernels. Concentrating on the foreground entity, we allocated 90\%\ of the rays for sampling from the foreground, with the residual 10\%\ designated for background sampling. In all our experiments, we utilize images of size 540x540. We employ the Adam optimizer~\cite{kingma2014adam} with the learning rate set to 1e-4 in training and it operates 10k iterations.
\subsection{Datasets}
\noindent \textbf{PeopleSnapshot}\ \ \ \ 
We utilize the PeopleSnapshot dataset~\cite{alldieck2018detailed}, containing recordings of individuals rotating before cameras which is available in its repository. For our experimental validation, we follow the protocol of Anim-NeRF~\cite{chen2021animatable}. Notably, the pose parameters derived using SMPLify~\cite{bogo2016keep} for this dataset do not consistently align with the images. Therefore, similar to Anim-NeRF, we optimize the human poses from the training and testing videos for our model and fix them during training to ensure a fair quantitative comparison. Moreover, our model is equipped to optimize body poses, a feature we exploit for all other results in the paper. For this purpose, we infer human pose parameters using HuMoR~\cite{rempe2021humor}, a state-of-the-art 3D human motion estimator. We refine these estimations through joint optimization within our model via back-propagation of the image reconstruction loss gradient on the pose parameters. The dataset includes camera parameters procured from a standardized calibration technique. \par

\vspace{1em}

\noindent\textbf{NeuMan dataset}\ \ \ \
As in NeuMan~\cite{jiang2022neuman}, we anticipate that our model's training data will consist of a video of a specific person conducting actions. Our model will use NeuMan's six videos: Seattle, Citron, Parking, Bike, Jogging, and Lab. They are available in NeuMan's repository. We will also use some of the videos in HuMoR~\cite{rempe2021humor} dataset, as NeuMan's videos do not contain occluded scenes. These videos are shot on cell phones and the scenes are in line with our daily use. Early research~\cite{mildenhall2021nerf, jiang2022neuman} shows that mobile phone videos are sufficient for NeRF network training. In addition, we will subsample the video frames, resulting in between 30 and 100 images depending on the length of the video. We will use COLMAP~\cite{schonberger2016structure} to derive camera positions, scene representation, and multi-view-stereo (MVS) depth maps from subsampled frames of the same scene. This follows the procedures established in NeuMan~\cite{jiang2022neuman} and other prior NeRF models~\cite{mildenhall2021nerf, wang2021neus, wang2023neus2}. In the data preprocessing phase of our study, we employed a structured approach to prepare the data for analysis. Initially, we utilized OpenPose~\cite{cao2017realtime} for the detection of 2D key points on human parts. Following the detection of 2D key points, we applied the HuMoR model~\cite{rempe2021humor} to combine these with OpenPose's 2D key points for the purpose of fitting the human body's SMPL (Skinned Multi-Person Linear model) estimation. This process enabled us to construct a three-dimensional representation of the human figure. Lastly, we utilized the SAM model~\cite{kirillov2023segment} for segmenting the human body from the background. This step was crucial for generating masks for training. \par
\subsection{Baselines}
\noindent\textbf{NeuMan}~\cite{jiang2022neuman}\ \ \ \
Two separate Neural Radiance Fields (NeRF) networks~\cite{mildenhall2021nerf} form the basis of the NeuMan system. NeuMan approximates the Skinned Multi-Person Linear (SMPL) body model~\cite{loper2015smpl} for the portrayal of humans. A rigid transformation is then used to translocate this approximation SMPL model into a standard or canonical coordinate framework. Additionally, using NeRF's processing capacity, the system uses this canonical space to directly determine the RGB colour spectrum and volumetric density of the human. NeuMan provides the capability to generate high-fidelity representations of human avatars and their environments from diverse and novel perspectives. We present outcomes for a fundamental setup of NeuMan, which operates across 500k iterations. \par
\vspace{1em}

\noindent\textbf{Instant-Avatar}~\cite{jiang2023instantavatar}\ \ \ \
InstantAvatar refines the standard neural radiance field methodology with an innovative variant to capture the canonical structure and visual attributes. It utilizes Instant-NGP~\cite{muller2022instant} to expedite the rendering of neural volumes for human avatar formation, employing a spatially efficient hash table as an alternative to traditional multilayer perceptions. We present outcomes for a fundamental setup of Instant-Avatar, which operates across 30k iterations.  \par
\vspace{1em}

\noindent\textbf{Anim-NeRF}~\cite{chen2021animatable}\ \ \ \
This foundational approach constructs human figures and their aesthetic features within a canonical space using an MLP-based Neural Radiance Field (NeRF). It initiates the process by generating an SMPL body configured to a specified pose. Subsequently, for each sample point in the deformed space, the corresponding skinning weights are determined by computing the weighted average of the nearest K vertices' skinning weights from the posed SMPL mesh. Ultimately, these skinning weights enable the transformation of the sample point back to the canonical space through the application of inverse Linear Blend Skinning (LBS). For Anim-NeRF, we configure its hyperparameters as follows: $|N(i)|$ = 4, $\delta$ = 0.2, $\lambda_1$ = 0.001, $\lambda_2$ = 0.01, and $\lambda_d$ = 0.1. It requires 200k iterations of training on the PeopleSnapshot Dataset. \par
\vspace{1em}

\noindent\textbf{Neural Body}~\cite{peng2021neural}\ \ \ \
This alternative baseline method concentrates on a spectrum of latent variables affixed to a malleable Skinned Multi-Person Linear (SMPL) mesh. These variables, anchored to the mesh, are decoded to generate pose-specific radiance fields. A neural network, predicated on these latent codes, regresses the density and color for each three-dimensional voxel. In the reconstruction phase, both the latent codes and the network are collaboratively learned from images spanning all video frames. For Neural Body, the configuration sets the initial learning rate at \(5 \times 10^{-4}\), with an exponential decay to \(5 \times 10^{-5}\) throughout the optimization process. Convergence on a four-view video consisting of 300 frames typically requires approximately 200k iterations. \par
\subsection{Comparison with SoTA 3D Pose Estimator}
\subsubsection*{Human Pose Estimation Comparison in the Normal Scene}
Prevailing Dynamic Human Neural Radiance Fields (NeRFs) typically rely on external 3D human pose estimators for predicting human posture, such as ROMP~\cite{sun2021monocular}. ROMP represents the forefront of single-frame SMPL~\cite{loper2015smpl} estimation within this domain. These estimators enable the capture and reasonable reconstruction of human motion parameters. We've improved our model for better motion capture, maintaining accurate human motion estimation even with occlusions, as seen with HuMoR~\cite{rempe2021humor}. Both models were tested against a video dataset featuring a range of movements, from simple gestures to complex actions, to validate their performance. These tests were conducted under uniform hardware and software conditions to maintain comparability. \par

\begin{figure}
    \centering
    \includegraphics[width=1\textwidth,height=0.39\textwidth]{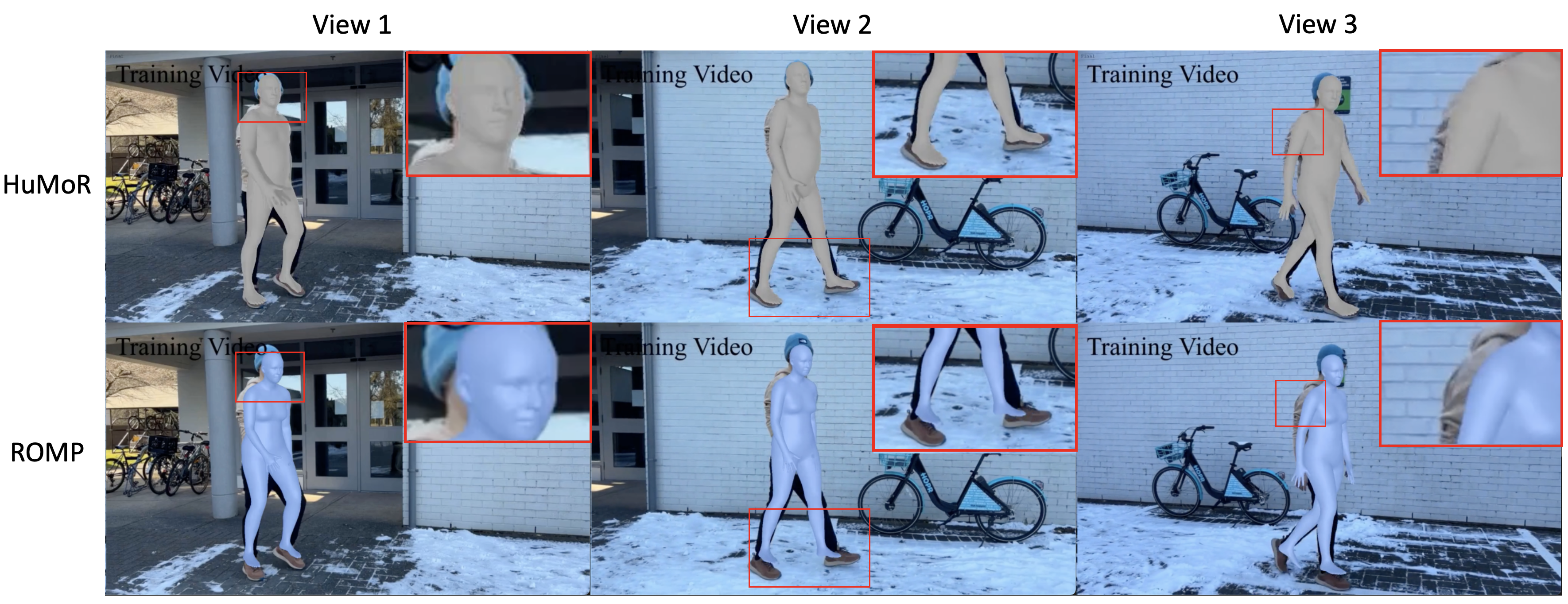}
    \caption{SMPL estimation results of HuMoR~\cite{rempe2021humor} (top) and ROMP~\cite{sun2021monocular} (bottom).}
    \label{fig:humorVSromp}
\end{figure}

The comparison in Figure \ref{fig:humorVSromp} highlights a marked improvement in SMPL parameter accuracy with the HuMoR model. HuMoR accurately estimates complex anatomical structures, notably in hands and heads, as shown by the mesh. This is further illustrated in figure \ref{fig:humorVSromp} \textbf{views 2} and \textbf{3}, where ROMP struggles with foot estimation in dynamic actions. HuMoR ensures consistent parameters for smoother SMPL estimation, in contrast to ROMP's jittery output. \par

\subsubsection*{Human Pose Estimation Comparison in the Occluded Scene}
\begin{figure}[htb]
    \centering
    \includegraphics[width=1\textwidth,height=0.39\textwidth]{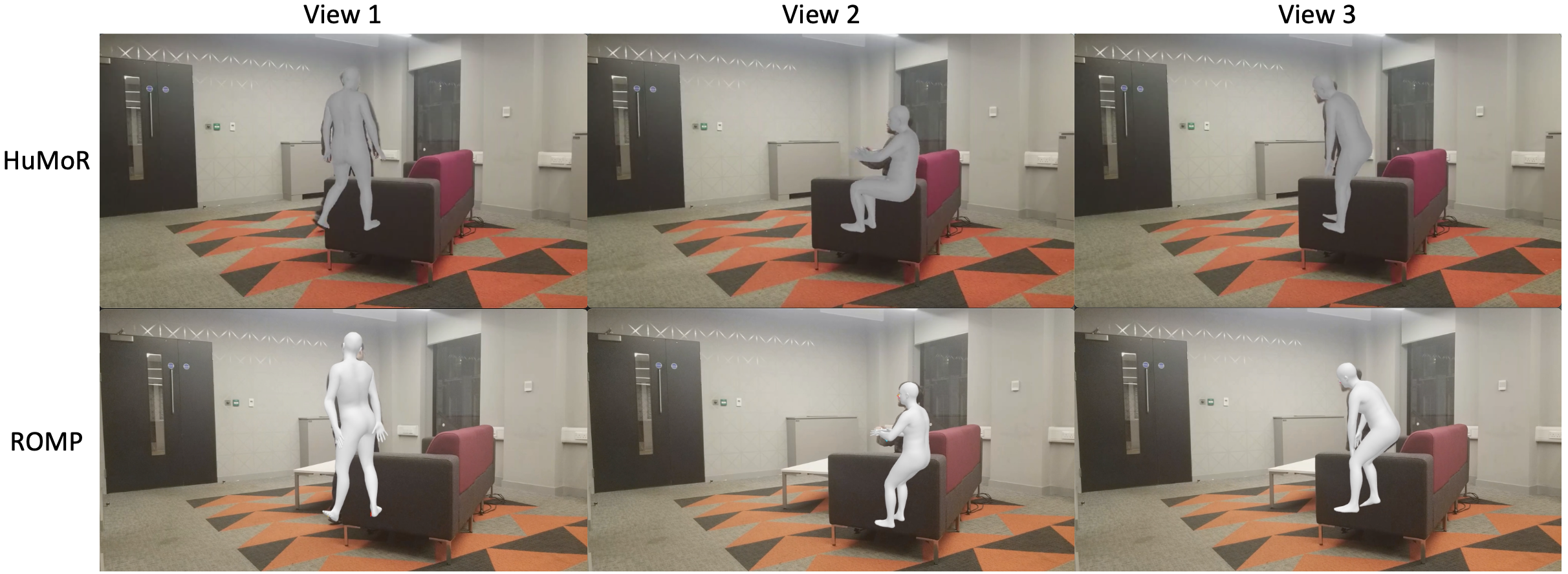}
    \caption{SMPL estimation results for HuMoR~\cite{rempe2021humor} (top) and ROMP~\cite{sun2021monocular} (bottom) in partially occluded scenes.}
    \label{fig:Occluded_Scene}
\end{figure}

Effective occlusion management is key in 3D reconstruction, with many models struggling to produce accurate SMPL parameter estimates. This challenge is highlighted in our extensive experiments comparing HuMoR and ROMP models. In occluded scenes, where the lower body of a subject is hidden, such as when seated on a sofa, our findings illustrate the models' divergent capabilities, as shown in Figure \ref{fig:Occluded_Scene}. HuMoR excels at estimating anatomically accurate SMPL meshes, leveraging data from adjacent frames to handle occlusions effectively. In contrast, ROMP's estimations often diverge from a realistic human shape. This is particularly evident in \textbf{view 1} of Figure \ref{fig:Occluded_Scene}, where it inaccurately predicts the positioning of the subject's legs. As demonstrated in \textbf{view 2}, where ROMP misinterprets the subject's posture, HuMoR's context-sensitive estimations diverge significantly from ROMP's error. Overall, our analyses confirm HuMoR's robustness and accuracy in occlusion scenarios, consistently outperforming ROMP by effectively utilizing temporal context. \par
\subsection{Comparison with SoTA Dynamic Human NeRFs}
\begin{figure}[!htbp]
    \centering
    \includegraphics[width=1\textwidth,height=0.38\textwidth]{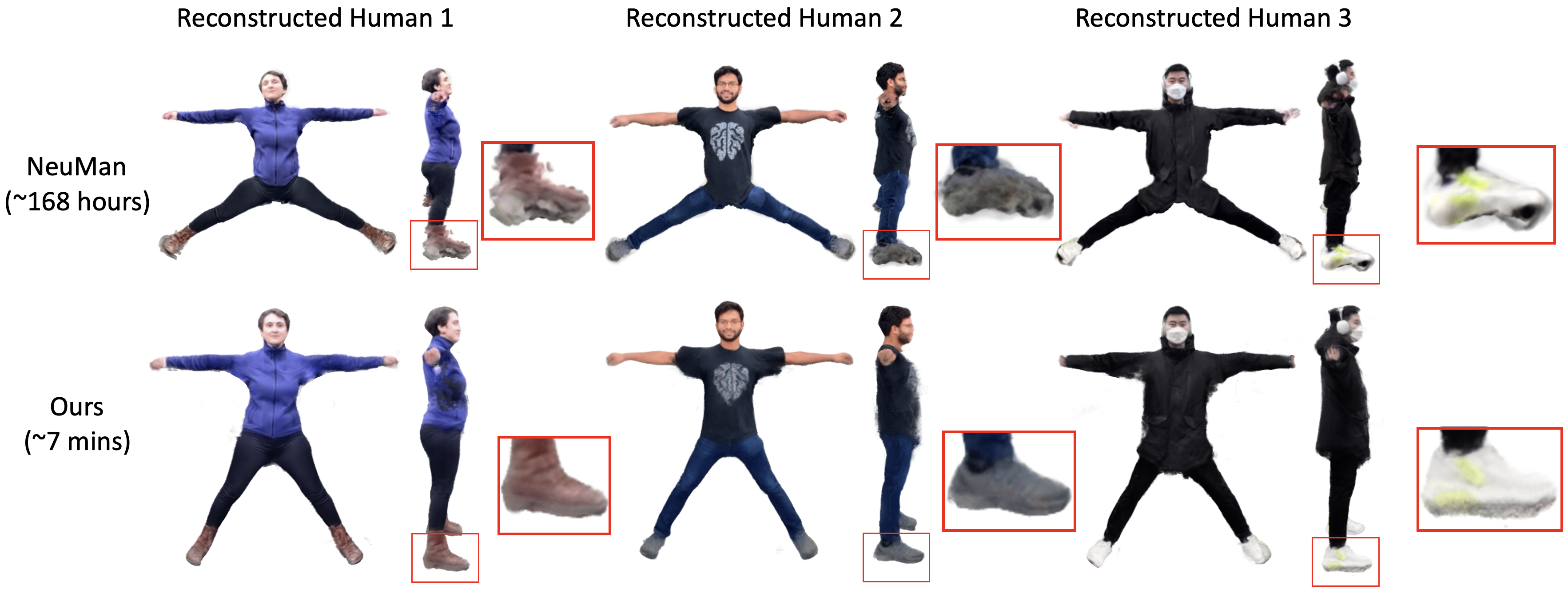}
    \caption{\textbf{Qualitative Results on NeuMan dataset.} We show front and side views of the reconstructed human avatar results in canonical space. Compared to SoTA NeuMan~\cite{jiang2022neuman}, our method converges much faster.}
    \label{fig:compare_neuman}
\end{figure}

\subsubsection{Experiment Results on NeuMan Dataset}
The NeuMan dataset contains six videos shot on a mobile phone and requires visual comparison for model evaluation due to the absence of quantitative metrics. The videos feature unbroken sequences of walking or running from the side, and turning motions, allowing for a complete capture of the body. Our model's partial reconstruction results are illustrated in Figure \ref{fig:compare_neuman}, compared against those from the NeuMan model~\cite{jiang2022neuman}. Figure \ref{fig:compare_neuman} shows that the NeuMan model, reliant on ROMP for SMPL~\cite{loper2015smpl} estimation, struggles with keeping poses consistent across frames. The step-by-step ROMP approach results in jarring transitions and visible mistakes, especially in dynamic areas like the feet which lead to blurred features such as the shoes. In contrast, our model results in a better reconstruction quality, especially in the feet area. While the NeRF-based human rendering in NeuMan is computationally intensive—requiring approximately 160 hours for the results shown in Figure \ref{fig:compare_neuman}—our model employs Instant-NGP~\cite{muller2022instant} rendering. This approach reduces the computation time to just 7 minutes and can be further decreased with smart frame sampling. Our approach not only achieves more precise and consistent avatar reconstructions but also significantly lower computational requirements. This results in making it an advantageous choice for 3D human avatar reconstruction tasks. \par

\begin{figure}[!htbp]
    \centering
    \includegraphics[width=1\textwidth,height=0.46\textwidth]{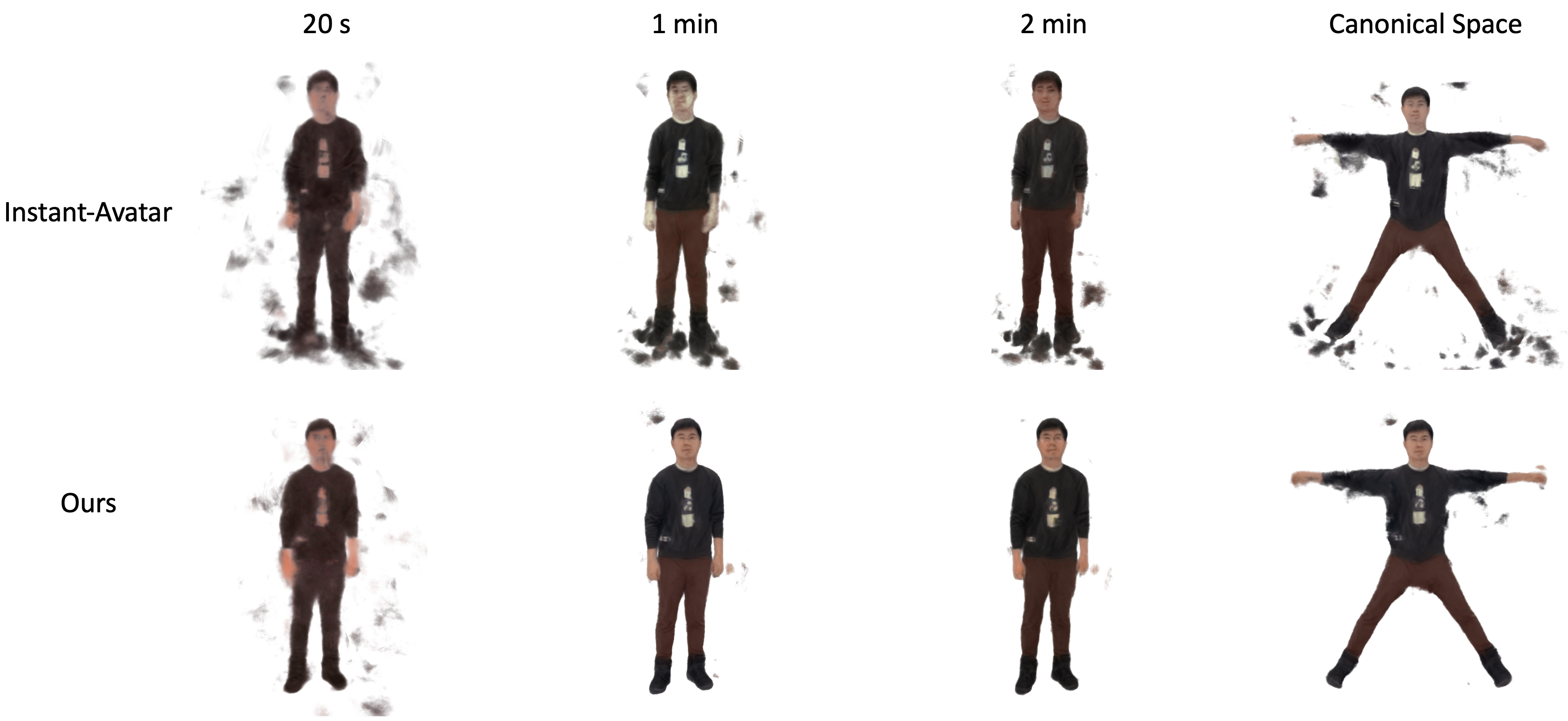}
    \caption{\textbf{Training Progression on NeuMan dataset.} We demonstrate the quality of the reconstructed human at various training iterations. Compared to SoTA Instant-Avatar~\cite{jiang2023instantavatar}, our method converges with higher quality and has less noise.}
    \label{fig:compare_avatar}
\end{figure}

In Figure \ref{fig:compare_avatar}, we showcase a detailed comparison of our model with Instant-Avatar~\cite{jiang2023instantavatar}. Both models utilize Instant-NGP~\cite{muller2022instant} for efficient rendering, but the core difference lies in the SMPL parameter estimation technique. In contrast, our model shows significantly fewer noisy pixels surrounding the human as a result of the intricate integration of a human motion prior, improving the reconstructions' precision and clarity. Eventually, the reconstructions become more refined, and our model generates accurate and distinct representations of the Da pose in canonical space. This comparison highlights the accuracy and resilience of our model both aesthetically and narratively, particularly in its capacity to reconstruct 3D human avatars devoid of environmental artefact interference.

\begin{figure}[htb]
    \centering
    \includegraphics[width=1\textwidth]{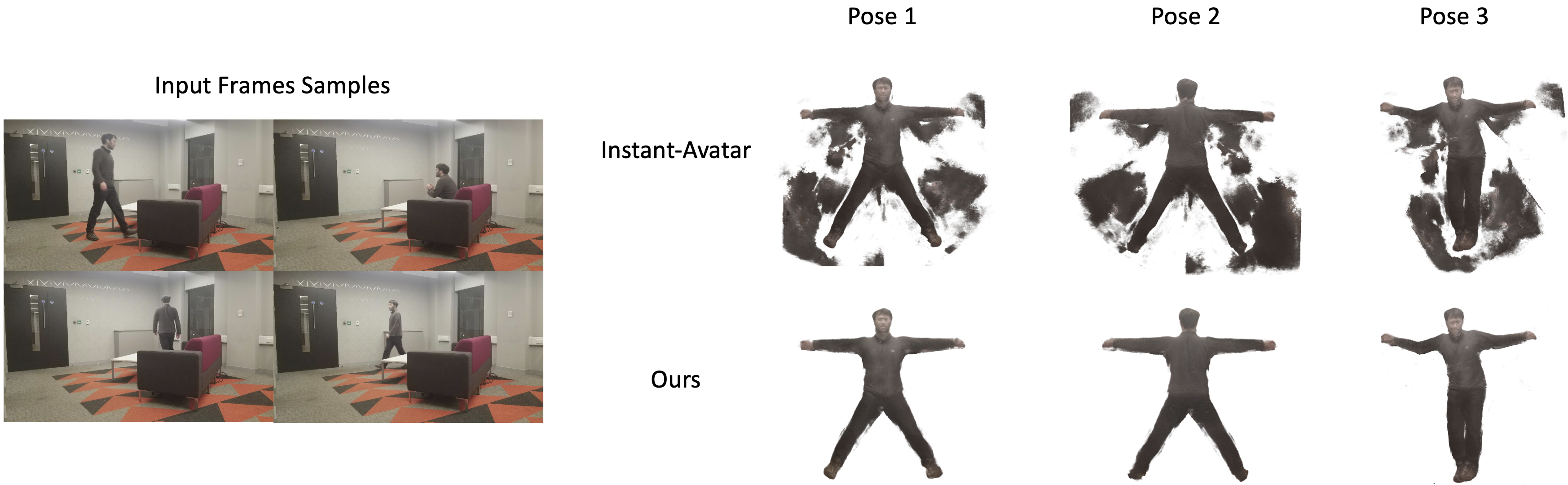}
    \caption{\textbf{Reconstruction Results in Occluded Scenes.} We present the novel pose as well as front and back views of the reconstructed human avatar results in canonical space. Compared to SoTA Instant-Avatar~\cite{jiang2023instantavatar}, our method converges with higher reconstruction quality in occluded scenes.}
    \label{fig:compare_avatar_occluded}
\end{figure}

\vspace{1em}

\noindent\textbf{Reconstruction Comparison in the Occluded Scene}\ \ \ \
\noindent Figure \ref{fig:compare_avatar_occluded} provides a visual comparison between our model and Instant-Avatar~\cite{jiang2023instantavatar}, specifically focusing on the challenge of reconstructing a human avatar in a scene where occlusion is present. In the training video, almost half of the human subject's lower body is obscured by a sofa, presenting a significant challenge for 3D reconstruction. The figure clearly showcases that our model is superior in reconstructing the occluded human figure, yielding a high-quality avatar. The reconstructed avatar from our model shows a remarkable reduction in extraneous pixels, resulting in a cleaner and more precise representation of the human form. Our framework is specifically designed to generate more accurate SMPL parameter predictions, even in scenarios involving occlusion or dynamic movement. The SMPL parameter estimation directly affects the sampling of human body parts during reconstruction. With our model's enhanced ability to closely approximate true human body configurations, the sampling process incurs fewer errors. Consequently, this leads to reconstructions with significantly better quality, characterized by fewer artifacts and noise. This indicates that our model has a robust understanding of human anatomy and can infer the occluded parts with greater accuracy. The visual fidelity of our avatar is not only more coherent with the visible portions but also maintains consistency in areas that are not directly observable. The performance of our model under these complex conditions suggests a sophisticated approach to interpreting and filling in unseen details. This likely involves leveraging a more advanced understanding of human shapes and postures. This feature is particularly important for applications that require realistic and accurate representation of human figures, even when those figures are partially obscured. \par
\subsubsection{Experiment Results on the PeopleSnapshot Dataset}

    \begin{table}[!htbp]
        \centering
        \scriptsize
	\begin{tabular}{C{3cm} C{3cm} C{3cm} C{3cm} C{3cm}}
	    \toprule 
            & female-3-casual & female-4-casual & male-3-casual & male-4-casual\\
            & \tiny{PSNR↑ SSIM↑ LPIPS↓} & \tiny{PSNR↑ SSIM↑ LPIPS↓} & \tiny{PSNR↑ SSIM↑ LPIPS↓} & \tiny{PSNR↑ SSIM↑ LPIPS↓}\\
		\midrule 
		Neural Body ($\sim$ 14 hours) &23.87 0.9504 0.0346&24.37 0.9451 0.0382&24.94 0.9428 0.0326&24.71 0.9469 0.0423\\
            Anim-NeRF ($\sim$ 13 hours) &\textbf{28.91} \textbf{0.9743} 0.0215&28.90 0.9678 0.0174&\textbf{29.37} 0.9703 0.0168&\textbf{28.37} 0.9605 \textbf{0.0268}\\
            Ours (5 minute)&28.08 0.9697 \textbf{0.0205}&\textbf{29.05} \textbf{0.9681} \textbf{0.0150}&29.27 \textbf{0.9707} \textbf{0.0155}&27.69 \textbf{0.9608} 0.0344\\
            \midrule 
            Anim-NeRF (5 minutes) &22.37 0.9311 0.0784&23.18 0.9292 0.0687&23.17 0.9266 0.0784&22.30 0.9235 0.0911 \\
            Ours (5 minutes)&\textbf{28.08 0.9697 0.0205}&\textbf{29.05 0.9681 0.0150}&\textbf{29.27 0.9707 0.0155}&\textbf{27.69 0.9608 0.0344}\\
            \midrule
            Anim-NeRF (1 minute) &11.71 0.7797 0.3321&12.31 0.8089 0.3344&12.39 0.7929 0.3393&13.10 0.7705 0.3460 \\
            Ours (1 minute) &\textbf{27.96 0.9763 0.0229}&\textbf{28.72 0.9635 0.0173}&\textbf{29.33 0.9779 0.0189}&\textbf{27.92 0.9656 0.0339} \\
		\bottomrule
	\end{tabular}
        \caption{\textbf{Qualitative Comparison with Anim-NeRF~\cite{chen2021animatable} and Neural Body~\cite{peng2021neural} on the PeopleSnapshot~\cite{alldieck2018detailed} dataset}. We compare actual pictures with images reconstructed by our approach and two dynamic human NeRF methods, Neural Body and Anim-NeRF, and give PSNR, SSIM, and LPIPS~\cite{zhang2018unreasonable} results. We evaluate Anim-NeRF and our technique with both 5 minutes and 1 minute of training time, as well as every approach at their convergence.}
        \label{tab:compare_anim}
    \end{table}

    \begin{figure}[htb]
        \centering
        \includegraphics[width=1\textwidth]{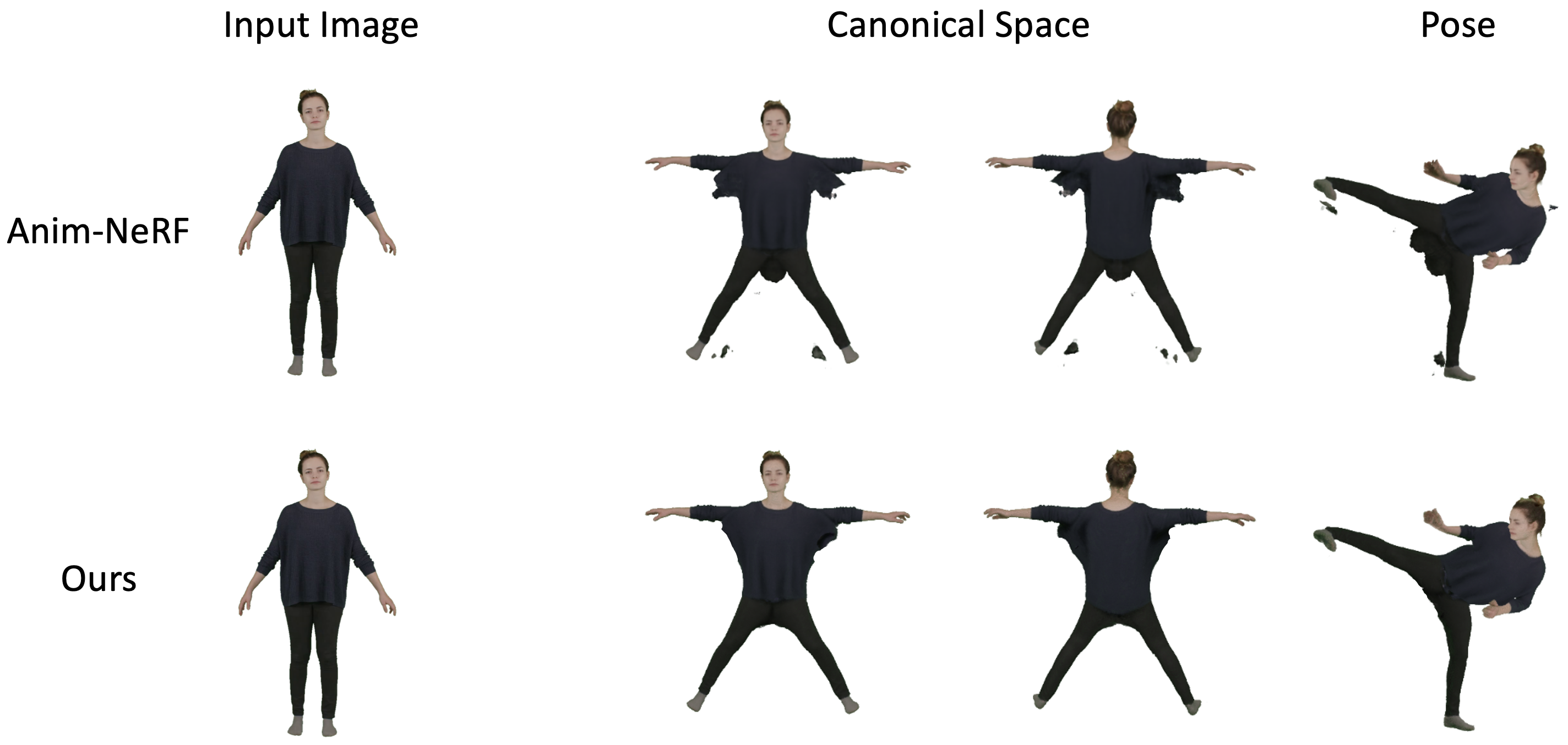}
        \caption{\textbf{Qualitative Results on PeopleSnapshot dataset~\cite{alldieck2018detailed}.} These datasets use a multi-camera system to capture the human and they provide the actual SMPL and real human avatars. We display reconstructed human avatars on PeopleSnapshot (top) in a variety of stances and from various vantage points. Compared to SoTA Anim-NeRF~\cite{jiang2023instantavatar}, our method converges with higher quality and a faster speed.}
        \label{fig:compare_anim}
    \end{figure}

Figure \ref{fig:compare_anim} and Table \ref{tab:compare_anim} from our most recent comparative analysis illustrate this intricate interplay by contrasting our model with the Anim-NeRF. PeopleSnapshot~\cite{alldieck2018detailed}, the datasets central to this comparison, are not simply ordinary; they exemplify meticulous curation and precision. PeopleSnapshot benefits from the accuracy of professional tools, unlike the NeuMan dataset, which was collected using mobile devices. Because of this distinction, the renderings produced using these datasets are of noticeably higher quality. Unlike mobile-captured datasets such as NeuMan, PeopleSnapshot is lab-created which offers accurate ground truth for SMPL values along with high-quality renderings. To evaluate the visual fidelity of avatar reconstructions, we use the PeopleSnapshot dataset for animation. The comparison information in Table \ref{tab:compare_anim} and the subjective assessments from Figure \ref{fig:compare_anim} support this. Anim-NeRF struggles with noise and inaccuracies even with its sophisticated approach, and our model stands out for its accurate and detailed renderings that faithfully capture the nuances of human anatomy. This is especially evident when rendering complex movements. The limited pose range of the PeopleSnapshot dataset does not adequately capture the quality of synthesising novel poses. Without actual ground truth for these new poses, we rely on a qualitative assessment, creating images of challenging new postures with our model and Anim-NeRF. Our method shines in these tests, handling complex positions with finesse and avoiding the artifacts seen with Anim-NeRF, such as distortions under the arms and between legs. Our model performs better, especially with loose clothing, because it is not limited by SMPL parameters for regularisation. To provide a basis for quantitative evaluation, we generate synthetic ground truth data for these difficult new poses. We demonstrate that our model not only outperforms Neural Body~\cite{peng2021neural} but also achieves parity with the state-of-the-art Anim-NeRF~\cite{chen2021animatable} in terms of quality. We base this comparison on the reconstructed images and their similarity to real images.

\vspace{1em}

\noindent\textbf{Speed}\ \ \ \
Compared to SoTA approaches, our method uses a lot less training time and computational resources. We only have to spend a couple of minutes to train on a single RTX 3090, but Anim- NeRF~\cite{chen2021animatable} needs two RTX 3090 trained for 13 hours, and Neural Body~\cite{peng2021neural} needs four RTX 2080 trained for 14 hours. We are able to produce reconstructed human avatars at 540$\times$540 resolution at a rate of greater than 15 frames per second with our framework, which is thousands of magnitude quicker than baselines. As results can be shown in Tab. \ref{tab:compare_anim}, our approach greatly outperforms Anim-NeRF in terms of picture quality for the same amount of training time. As demonstrated in Fig. \ref{fig:compare_avatar}, our training process significantly outperforms Instant-Avatar's. Our system captures considerable details and achieves satisfactory visual quality within just 20 seconds, with further enhancements in 30 seconds. Our approach achieves high-fidelity reconstruction within just a minute of training, while Instant-Avatar produces lower-quality outcomes at this early stage. \par
\subsection{Ablation Study}
\subsubsection{Posture-sensitive Space Reduction}
    
    \begin{table}[!htbp]
        \centering
        \footnotesize
	\begin{tabular}{C{3cm} C{3cm} C{3cm} C{3cm} C{3cm}}
	    \toprule 
		&PSNR↑& SSIM↑& LPIPS↓& Training Time↓\\
            \midrule
            w/o Skipping&27.92&0.9720&0.029&3m 11s\\
            w/ Skipping&28.10&0.9713&0.031&1m 57s\\
            \midrule
            Decay Rate=0.6&28.02&0.9514&0.031&1m 49s\\
            Decay Rate=0.75&28.23&0.9719&0.032&1m 41s\\
		\bottomrule
	\end{tabular}
        \caption{\textbf{Ablation Study: Posture-sensitive Space Reduction Approach.} Regarding the hyperparameters of the posture-sensitive space reduction method, we conduct an ablation analysis. In PeopleSnapshot dataset after fifty epochs, we offer the mean of 4 sequences for each trial.}
        \label{tab:ablation_empty}
    \end{table}
    
\noindent In our model, we use a technique called posture-sensitive space reduction to reduce redundant samplings during the rendering procedure. For dynamic humans, we examine the impact of our posture-sensitive space reduction approach. The training and rendering performances are greatly improved by avoiding empty space, as shown in Tab. \ref{tab:ablation_empty}, and this effect is independent of the hyperparameter selection. \par
    \subsubsection{Occupancy-based Regularization Loss Term}
    
    \begin{table}[!htbp]
        \centering
        \footnotesize
    	\begin{tabular}{C{3cm} C{3cm} C{3cm} C{3cm}}
    	    \toprule 
                &PSNR↑& SSIM↑& LPIPS↓\\
                \midrule
                w/o occupancy-based term&27.94&0.9590&0.0342\\
                w occupancy-based term&28.31&0.9630&0.0230\\
    		\bottomrule
    	\end{tabular}
        \caption{\textbf{Ablation Study: Occupancy-based Regularization Loss Term.} We assess the quality of images as an average over the four PeopleSnapshot training sets.}
        \label{tab:ablation_loss}
    \end{table}

\noindent By using our regularisation loss $L_{\text{reg}}$ from Section 4.6, the occupancy field for posture-sensitive space reduction may also assist in regularising the radiance field and decreasing noise. As shown by the PSNR increase in Tab. \ref{tab:ablation_loss}, this loss successfully lowers floating artefacts and thus aids in improving the overall image quality. Encouraging zero density at all points in space is a different typical strategy for decreasing floating noise. We contrast this approach with our own and discover that it (Global Sparsity) results in a worse quality. \par
\subsection{Limitation}
\begin{figure}[htb]
    \centering
    \includegraphics[width=0.5\textwidth]{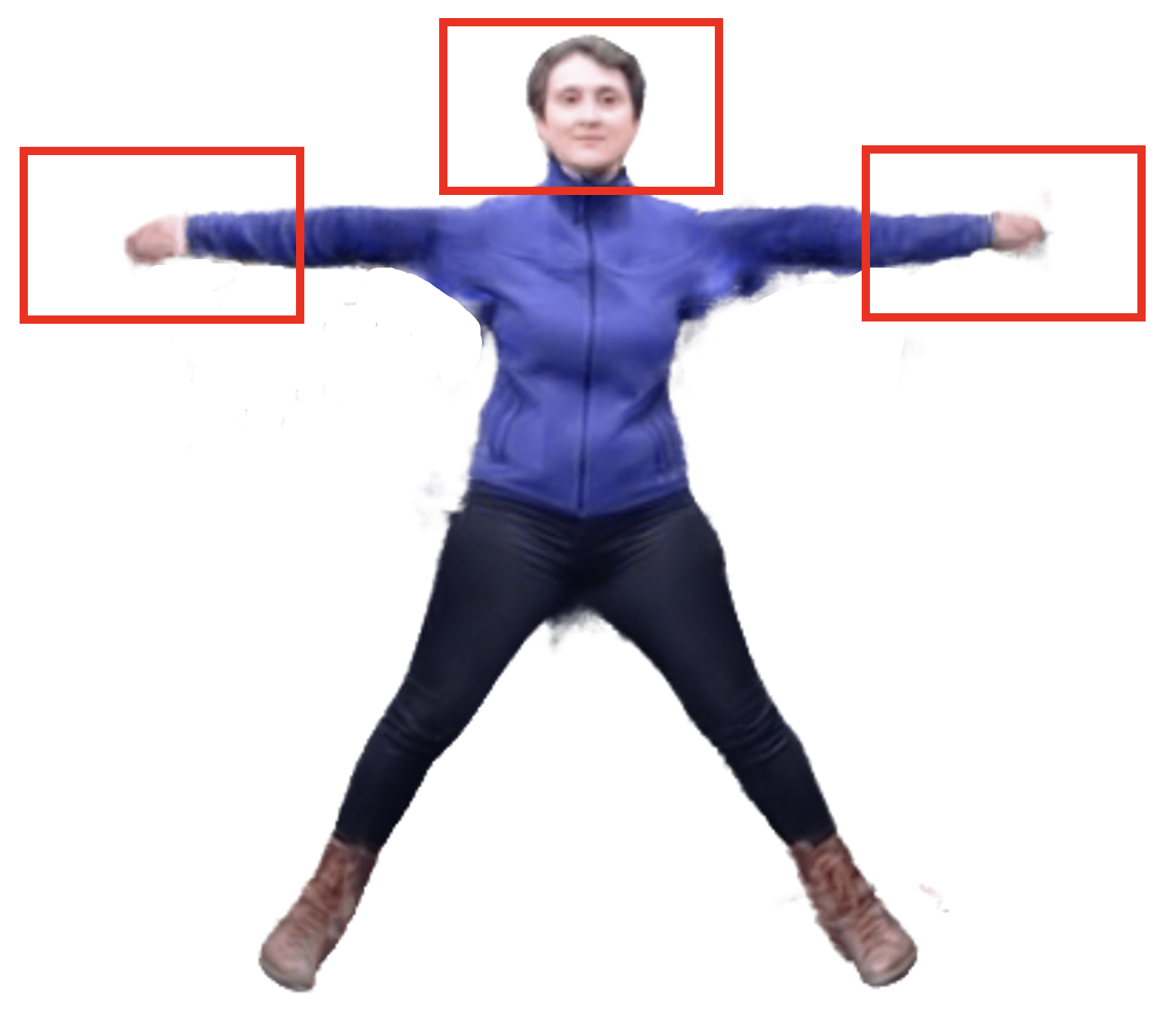}
    \caption{\textbf{Limitation: Cannot model facial expressions and the details of hands.}}
    \label{fig:limitation1}
\end{figure}

\noindent
Figure \ref{fig:limitation1} illustrates a limitation in our model: the nuanced capture of facial expressions and the fine details of hands. Even with the HuMoR model improving SMPL estimation accuracy, our model still faces challenges in accurately capturing the complex movements of facial muscles and the nuances of hand gestures. This difficulty underscores the complexity of human expressions and gestures, composed of subtle signals and rapid changes. Future work could further enhance the quality of our reconstructions by incorporating models such as SMPL-H~\cite{romero2022embodied} or SMPL-X~\cite{pavlakos2019expressive}. These models provide a more refined representation of facial and hand details, potentially elevating the quality of our reconstructions. \par

%

A crucial element of our conversation concerns the computational complexity entailed in rendering and reconstructing human figures. The precise calculation of pixel values is essential to this process. This intensity results from the need to accurately calculate light's interactions with the human model across a large number of frames. The computational demands increase significantly as the reconstruction's resolution increases. This can be mainly ascribed to the exponential rise in the number of pixels that need to be processed individually. Consequently, longer training times and more computational power are required to achieve higher-resolution reconstructions.

There are limitations to our model's reliance on visual data, particularly when attempting to infer details from areas that are obscured or invisible. For instance, it's still difficult to reconstruct a subject's back from a video that only shows their front. A solution might be found by expanding the model to include learning-based techniques, which would allow for the prediction and full reconstruction of these invisible areas for an avatar that is fully realised. \par
\section{Conclusion}
In this paper, we provide an approach that can quickly create animateable human avatars from motion sequences and then generate and display the model at 15 frames per second. We integrate the human motion model HuMoR with the effective neural radiance field Instant-NGP~\cite{muller2022instant} and the effective articulated model Fast-SNARF~\cite{chen2023fast} to accomplish it. The speed of this simple combination is not ideal. To increase rendering performance and decrease drifting artefacts in the area, we designed an SMPL-based optimization for human avatar, occupancy-based regularisation loss and a posture-sensitive space reduction strategy. Our approach delivers comparable picture quality while being noticeably quicker during inference and learning than the SoTA NeRF-based approaches. Although the focus of this study is human reconstruction, the concept is transferrable to other issues. The intriguing next stage is to expand our methodology to produce human geometry mesh and human avatars with efficient generation of hand and facial information.
\newpage
\bibliographystyle{IEEEtran}
\bibliography{IEEEabrv, main}
\end{document}